\documentclass[sigconf]{acmart}
\usepackage{booktabs} 
\usepackage{multirow}
\usepackage{color}
\usepackage{subfigure} 
\usepackage{array}
\usepackage{breqn}
\usepackage{{inputenc}}
\usepackage{balance}
\usepackage{multirow,tabularx}
\usepackage{longtable}
\usepackage{booktabs,longtable}
\usepackage{hhline}
\usepackage[flushleft]{threeparttable}
\usepackage{algorithm}
\usepackage{algorithmic}
\usepackage{epsfig}
\usepackage{graphicx}
\usepackage{epstopdf}
\usepackage{thmtools}
\usepackage{enumitem}
\usepackage{verbatim}
\usepackage{amsfonts}
\usepackage{hyperref}
\hypersetup{colorlinks=false,linkcolor=blue,urlcolor=blue,citecolor=red}
\newcommand\norm[1]{\left\lVert#1\right\rVert}
\newcommand{\Lapl}{\mathbf{\mathop{\mathcal{L}}}}
\newcommand{\Trans}[1]{{#1}^{\top}}

\newcommand{\Mat}[1]{\mathbf{#1}}

\newcommand{\Space}[1]{\mathbb{#1}}
\newcommand{\Set}[1]{\mathcal{#1}}

\newcommand{\ie}{\emph{i.e., }}
\newcommand{\eg}{\emph{e.g., }}

\newcommand{\etc}{\emph{etc.}}
\newcommand{\wrt}{\emph{w.r.t. }}
\newcommand{\cf}{\emph{cf. }}
\newcommand{\aka}{\emph{aka. }}

\theoremstyle{definition}

\def\BibTeX{{\rm B\kern-.05em{\sc i\kern-.025em b}\kern-.08emT\kern-.1667em\lower.7ex\hbox{E}\kern-.125emX}}

\begin{document}

\settopmatter{printacmref=true}
\fancyhead{}

\title{KGAT: Knowledge Graph Attention Network for Recommendation}

\author{Xiang Wang}
\affiliation{%
	\institution{National University of Singapore}
}
\email{xiangwang@u.nus.edu}

\author{Xiangnan He}
\authornote{Xiangnan He is the corresponding author.}
\affiliation{%
	\institution{University of Science and Technology of China}
}
\email{xiangnanhe@gmail.com}

\author{Yixin Cao}
\affiliation{%
	\institution{National University of Singapore}
}
\email{caoyixin2011@gmail.com}

\author{Meng Liu}
\affiliation{%
 \institution{Shandong University}
}
\email{mengliu.sdu@gmail.com}

\author{Tat-Seng Chua}
\affiliation{%
	\institution{National University of Singapore}
}
\email{dcscts@nus.edu.sg}

\begin{abstract}

To provide more accurate, diverse, and explainable recommendation, it is compulsory to go beyond modeling user-item interactions and take side information into account. Traditional methods like factorization machine (FM) cast it as a supervised learning problem, which assumes each interaction as an independent instance with side information encoded. Due to the overlook of the relations among instances or items (\eg the director of a movie is also an actor of another movie), these methods are insufficient to distill the collaborative signal from the collective behaviors of users. 

In this work, we investigate the utility of knowledge graph (KG), which breaks down the independent interaction assumption by linking items with their attributes. We argue that in such a hybrid structure of KG and user-item graph, high-order relations --- which connect two items with one or multiple linked attributes --- are an essential factor for successful recommendation. We propose a new method named \textit{Knowledge Graph Attention Network} (KGAT) which explicitly models the high-order connectivities in KG in an end-to-end fashion. It recursively propagates the embeddings from a node's neighbors (which can be users, items, or attributes) to refine the node's embedding, and employs an attention mechanism to discriminate the importance of the neighbors. Our KGAT is conceptually advantageous to existing KG-based recommendation methods, which either exploit high-order relations by extracting paths or implicitly modeling them with regularization. Empirical results on three public benchmarks show that KGAT significantly outperforms state-of-the-art methods like Neural FM~\cite{NFM} and RippleNet~\cite{RippleNet}. Further studies verify the efficacy of embedding propagation for high-order relation modeling and the interpretability benefits brought by the attention mechanism. We release the codes and datasets at \url{https://github.com/xiangwang1223/knowledge_graph_attention_network}.

\end{abstract}

\copyrightyear{2019}
\acmYear{2019}
\setcopyright{acmcopyright}
\acmConference[KDD '19]{The 25th ACM SIGKDD Conference on Knowledge
Discovery and Data Mining}{August 4--8, 2019}{Anchorage, AK, USA}
\acmBooktitle{The 25th ACM SIGKDD Conference on Knowledge Discovery and Data
Mining (KDD '19), August 4--8, 2019, Anchorage, AK, USA}
\acmPrice{15.00}
\acmDOI{10.1145/3292500.3330989}
\acmISBN{978-1-4503-6201-6/19/08}

%
%

\maketitle

\section{Introduction}
The success of recommendation system makes it prevalent in Web applications, ranging from search engines, E-commerce, to social media sites and news portals --- without exaggeration, almost every service that provides content to users is equipped with a recommendation system. 
To predict user preference from the key (and widely available) source of user behavior data, much research effort has been devoted to collaborative filtering (CF)~\cite{NAIS,NCF,NGCF}. Despite its effectiveness and universality, CF methods suffer from the inability of modeling side information~\cite{NSCR,TEM}, such as item attributes, user profiles, and contexts, thus perform poorly in sparse situations where users and items have few interactions. 
To integrate such information, a common paradigm is to transform them into a generic feature vector, together with user ID and item ID, and feed them into a supervised learning (SL) model to predict the score. Such a SL paradigm for recommendation has been widely deployed 
in industry~\cite{DIN,DeepCrossing,WideDeep}, and some representative models include factorization machine (FM)~\cite{FM}, NFM (neural FM)~\cite{NFM}, Wide\&Deep~\cite{WideDeep}, and xDeepFM~\cite{xDeepFM}, \etc 

\begin{figure}[t]
    \centering
	\includegraphics[width=0.4\textwidth]{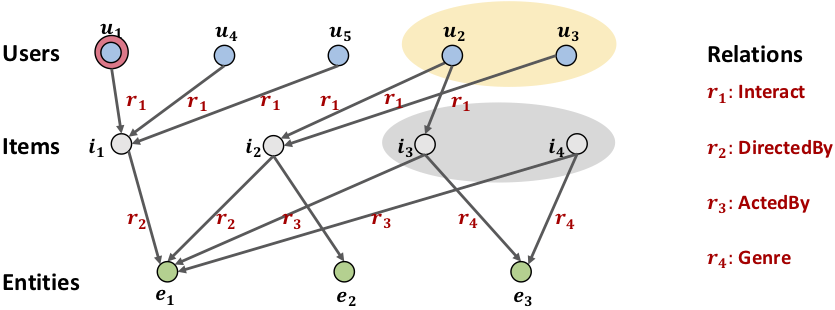}
	\vspace{-10pt}
	\caption{A toy example of collaborative knowledge graph. $u_1$ is the target user to provide recommendation for. The yellow circle and grey circle denote the important users and items discovered by high-order relations but are overlooked by traditional methods. Best view in color.
	}
	\label{fig:intro}
	\vspace{-15pt}
\end{figure}


Although these methods have provided strong performance, a deficiency is that they model each interaction as an independent data instance and do not consider their relations. 
This makes them insufficient to distill attribute-based collaborative signal from the collective behaviors of users.
As shown in Figure \ref{fig:intro}, there is an interaction between user $u_1$ and movie $i_1$, which is \textit{directed by} the person $e_1$. CF methods focus on the histories of similar users who also watched $i_1$, \ie $u_4$ and $u_5$; while SL methods emphasize the similar items with the attribute $e_1$, \ie $i_2$. Obviously, these two types of information not only are complementary for recommendation, but also form a high-order relationship between a target user and item together. However, existing SL methods fail to unify them and cannot take into account the high-order connectivity, such as the users in the yellow circle who watched \textbf{other movies} directed by the same person $e_1$, or the items in the grey circle that share \textbf{other common relations} with $e_1$.

To address the limitation of feature-based SL models, a solution is to take the graph of item side information, \aka knowledge graph\footnote{A KG is typically described as a heterogeneous network consisting of entity-relation-entity triplets, where the entity can be an item or an attribute.}~\cite{cao2018joint,cao2018neural}, into account to construct the predictive model. We term the hybrid structure of knowledge graph and user-item graph as \textit{collaborative knowledge graph} (CKG).
As illustrated in Figure \ref{fig:intro}, the key to successful recommendation is to fully exploit the high-order relations in CKG, \eg the long-range connectivities:
\begin{itemize}[leftmargin=*]
    \item $u_1\xrightarrow{r_1}i_1\xrightarrow{-r_2} e_1 \xrightarrow{r_2} i_2 \xrightarrow{-r_1} \{u_2, u_3\},$
    \item $u_1\xrightarrow{r_1}i_1\xrightarrow{-r_2} e_1 \xrightarrow{r_3}\{i_3, i_4\},$
\end{itemize}
which represent the way to the yellow and grey circle, respectively.
Nevertheless, to exploit such high-order information the challenges are non-negligible: 1) the nodes that have high-order relations with the target user increase dramatically with the order size, which imposes computational overload to the model, and 2) the high-order relations contribute unequally to a prediction, which requires the model to carefully weight (or select) them. 

Several recent efforts have attempted to leverage the CKG structure for recommendation, which can be roughly categorized into two types, path-based~\cite{DBLP:conf/wsdm/YuRSGSKNH14,MetaFMG,MCRec,KGreasoning19,KGRnn18,RippleNet} and regularization-based~\cite{CKE,KGreasoning19,KGMemory18,KTUP}: 
\begin{itemize}[leftmargin=*]
    \item Path-based methods extract paths that carry the high-order information and feed them into predictive model. To handle the large number of paths between two nodes, they have either applied path selection algorithm 
    to select prominent paths~\cite{KGRnn18,KGreasoning19}, 
    or defined meta-path patterns to constrain the paths~\cite{yu2013collaborative,MCRec}. One issue with such two-stage methods is that the first stage of path selection has a large impact on the final performance, but it is not optimized for the recommendation objective. Moreover, defining effective meta-paths requires domain knowledge, which can be rather labor-intensive for complicated KG with diverse types of relations and entities, since many meta-paths have to be defined to retain model fidelity.
    \item Regularization-based methods devise additional loss terms that capture the KG structure to regularize the recommender model learning. 
    For example, KTUP~\cite{KTUP} and CFKG~\cite{CFKG} jointly train the two tasks of recommendation and KG completion with shared item embeddings. 
    Instead of directly plugging high-order relations into the model optimized for recommendation, these methods only encode them in an implicit manner. Due to the lack of an explicit modeling, neither the long-range connectivities are guaranteed to be captured, nor the results of high-order modeling are interpretable. 
\end{itemize}

Considering the limitations of existing solutions, we believe it is of critical importance to develop a model that can exploit high-order information in KG in an efficient, explicit, and end-to-end manner. Towards this end, we take inspiration from the recent developments of 
graph neural networks~\cite{GCN,GraphSage,GAT}, which have the potential of achieving the goal but have not been explored much for KG-based recommendation. 
Specifically, we propose a new method named \textit{Knowledge Graph Attention Network} (KGAT), which is equipped with two designs to correspondingly address the challenges in high-order relation modeling:
1) recursive embedding propagation, which updates a node's embedding based on the embeddings of its neighbors, and recursively performs such embedding propagation to capture high-order connectivities in a linear time complexity; and 
2) attention-based aggregation, which employs the neural attention mechanism~\cite{Attention,ACF} to learn the weight of each neighbor during a propagation, such that the attention weights of cascaded propagations can reveal the importance of a high-order connectivity.
Our KGAT is conceptually advantageous to existing methods in that: 1) compared with path-based methods, it avoids the labor-intensive process of materializing paths, thus is more efficient and convenient to use, and 2) compared with regularization-based methods, it directly factors high-order relations into the predictive model, thus all related parameters are tailored for optimizing the recommendation objective. 

The contributions of this work are summarized as follows:
\begin{itemize}[leftmargin=*]
    \item We highlight the importance of explicitly modeling the high-order relations in collaborative knowledge graph to provide better recommendation with item side information. 
    \item We develop a new method KGAT, which achieves high-order relation modeling in an explicit and end-to-end manner under the graph neural network framework. 
    \item We conduct extensive experiments on three public benchmarks, demonstrating the effectiveness of KGAT and its interpretability in understanding the importance of high-order relations. 

\end{itemize}

\section{Task Formulation}

\begin{figure*}[ht]
    \centering
	\includegraphics[width=0.85\textwidth]{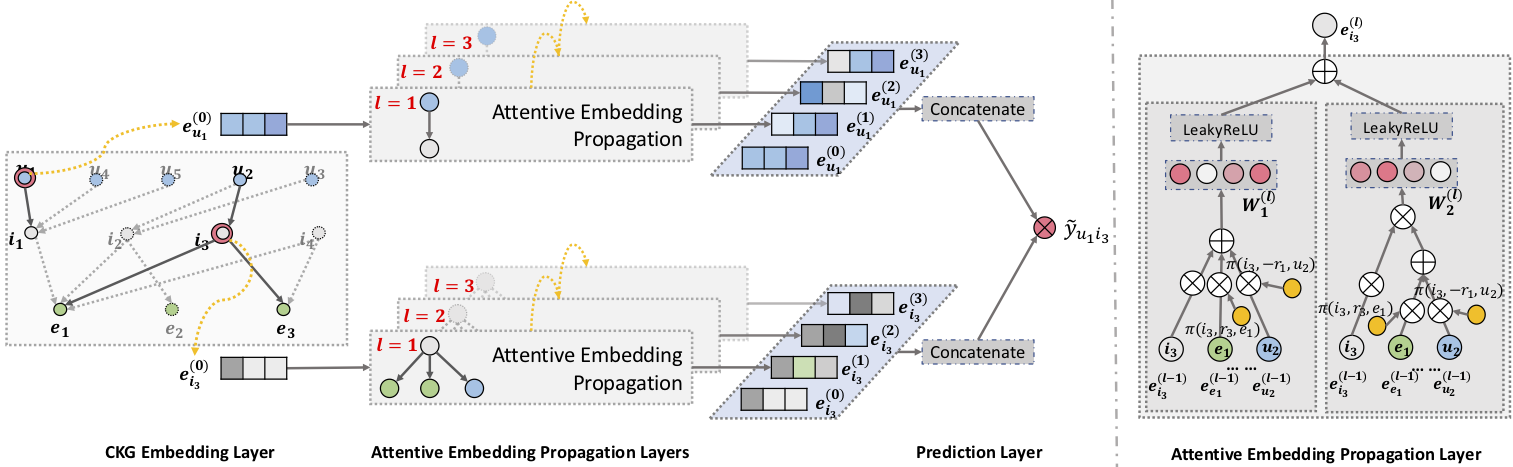}
	\vspace{-10pt}
	\caption{Illustration of the proposed KGAT model. The left subfigure shows model framework of KGAT, and the right subfigure presents the attentive embedding propagation layer of KGAT.}
	\label{fig:framework}
	\vspace{-10pt}
\end{figure*}

We first introduce the concept of CKG and highlight the high-order connectivity among nodes, as well as the compositional relations.

\vspace{5px}
\noindent\textbf{User-Item Bipartite Graph:}
In a recommendation scenario, we typically have historical user-item interactions (\eg purchases and clicks).
Here we represent interaction data as a user-item bipartite graph $\Set{G}_1$, which is defined as $\{(u,y_{ui},i)|u\in\Set{U},i\in\Set{I})\}$, where $\Set{U}$ and $\Set{I}$ separately denote the user and item sets, and a link $y_{ui}=1$ indicates that there is an observed interaction between user $u$ and item $i$; otherwise $y_{ui}=0$.

\vspace{5px}
\noindent\textbf{Knowledge Graph}.
In addition to the interactions, we have side information for items (\eg item attributes and external knowledge).
Typically, such auxiliary data consists of real-world entities and relationships among them to profile an item.
For example, a movie can be described by its director, cast, and genres.
We organize the side information in the form of knowledge graph $\Set{G}_2$, which is a directed graph composed of subject-property-object triple facts~\cite{KTUP}.
Formally, it is presented as $\{(h,r,t)|h,t\in\Set{E},r\in\Set{R}\}$, where each triplet describes that there is a relationship $r$ from head entity $h$ to tail entity $t$.
For example, (\emph{Hugh Jackman, ActorOf, Logan}) states the fact that Hugh Jackman is an actor of the movie Logan.
Note that $\Set{R}$ contains relations in both canonical direction (\eg \emph{ActorOf}) and inverse direction (\eg \emph{ActedBy}).
Moreover, we establish a set of item-entity alignments $\Set{A}=\{(i,e)|i\in\Set{I},e\in\Set{E}\}$, where $(i,e)$ indicates that item $i$ can be aligned with an entity $e$ in the KG.





\vspace{5px}
\noindent\textbf{Collaborative Knowledge Graph}.
Here we define the concept of CKG, which encodes user behaviors and item knowledge as a unified relational graph.
We first represent each user behavior as a triplet, $(u,\emph{Interact},i)$, where $y_{ui}=1$ is represented as an additional relation \emph{Interact} between user $u$ and item $i$.
Then based on the item-entity alignment set, the user-item graph can be seamlessly integrated with KG as a unified graph $\Set{G}=\{(h,r,t)|h,t\in\Set{E}',r\in\Set{R}'\}$, where $\Set{E}'=\Set{E}\cup\Set{U}$ and $\Set{R}'=\Set{R}\cup\{\emph{Interact}\}$.

\vspace{5px}
\noindent\textbf{Task Description}
We now formulate the recommendation task to be addressed in this paper:
\begin{itemize}[leftmargin=*]
    \item \textbf{Input}: collaborative knowledge graph $\Set{G}$ that includes the user-item bipartite graph $\Set{G}_1$ and knowledge graph $\Set{G}_2$.
    \item \textbf{Output}: a prediction function that predicts the probability $\hat{y}_{ui}$ that user $u$ would adopt item $i$.
\end{itemize}




\vspace{5px}
\noindent\textbf{High-Order Connectivity}.
Exploiting high-order connectivity is of importance to perform high-quality recommendation.
Formally, we define the \emph{$L$-order connectivity} between nodes as a multi-hop relation path: $e_{0}\xrightarrow{r_{1}}e_{1}\xrightarrow{r_{2}}\cdots\xrightarrow{r_{L}}e_{L}$, where $e_{l}\in\Set{E}'$ and $r_{l}\in\Set{R}'$;
$(e_{l-1},r_{l},e_{l})$ is the $l$-th triplet, and $L$ is the length of the sequence.
To infer user preference, CF methods build upon behavior similarity among users --- more specifically similar users would exhibit similar preferences on items.
Such intuition can be represented as behavior-based connectivity like $u_1\xrightarrow{r_1} i_{1}\xrightarrow{-r_1} u_2\xrightarrow{r_1} i_2$,
which suggests that $u_1$ would exhibit preference on $i_2$, since her similar user $u_2$ has adopted $i_2$ before.
Distinct from CF methods, SL models like FM and NFM focus on attributed-based connectivity, assuming that users tend to adopt items that share similar properties.
For example, $u_1\xrightarrow{r_1} i_{1}\xrightarrow{r_2} e_1 \xrightarrow{-r_2} i_2$ suggests that $u_1$ would adopt $i_2$ since it has the same director $e_1$ with $i_1$ she liked before.
However, FM and NFM treat entities as the values of individual feature fields, failing to reveal relatedness across fields and related instances.
For instance, it is hard to model $u_1\xrightarrow{r_1} i_{1}\xrightarrow{r_2} e_1 \xrightarrow{-r_3} i_2$, although $e_1$ serves as the bridge connecting \emph{director} and \emph{actor} fields.
We therefore argue that these methods do not fully explore the high-order connectivity and leave compositional high-order relations untouched.

\section{Methodology}

We now present the proposed KGAT model, which exploits high-order relations in an end-to-end fashion.
Figure~\ref{fig:framework} shows the model framework, which consists of three main components:
1)~embedding layer, which parameterizes each node as a vector by preserving the structure of CKG;
2)~attentive embedding propagation layers, which recursively propagate embeddings from a node's neighbors to update its representation, and employ knowledge-aware attention mechanism to learn the weight of each neighbor during a propagation;
and 3)~prediction layer, which aggregates the representations of a user and an item from all propagation layers, and outputs the predicted matching score.


\subsection{Embedding Layer}\label{sec:embedding-propagation}

Knowledge graph embedding is an effective way to parameterize entities and relations as vector representations, while preserving the graph structure.
Here we employ TransR~\cite{TransR}, a widely used method, on CKG.
To be more specific, it learns embeds each entity and relation by optimizing the translation principle $\Mat{e}^{r}_{h}+\Mat{e}_{r}\approx \Mat{e}^{r}_{t}$, if a triplet $(h,r,t)$ exists in the graph.
Herein, $\Mat{e}_{h},\Mat{e}_{t}\in\Space{R}^{d}$ and $\Mat{e}_{r}\in\Space{R}^{k}$ are the embedding for $h$, $t$, and $r$, respectively;
and $\Mat{e}_{h}^{r},\Mat{e}_{t}^{r}$ are the projected representations of $\Mat{e}_{h}$ and $\Mat{e}_{t}$ in the relation $r$'s space. 
Hence, for a given triplet $(h,r,t)$, its plausibility score (or energy score) is formulated as follows:
\begin{align}\label{transr}
    g(h,r,t)=\norm{\Mat{W}_{r}\Mat{e}_{h}+\Mat{e}_{r} - \Mat{W}_{r}\Mat{e}_{t}}_{2}^{2},
\end{align}
where $\Mat{W}_{r}\in\Space{R}^{k\times d}$ is the transformation matrix of relation $r$, which projects entities from the $d$-dimension entity space into the $k$-dimension relation space.
A lower score of $g(h,r,t)$ suggests that the triplet is more likely to be true true, and vice versa.

The training of TransR considers the relative order between valid triplets and broken ones, and encourages their discrimination through a pairwise ranking loss:
\begin{align}\label{equ:kg-loss}
    \Lapl_{\text{KG}}=\sum_{(h,r,t,t')\in\Set{T}}-\ln\sigma\Big(g(h,r,t')-g(h,r,t)\Big),
\end{align}
where $\Set{T}=\{(h,r,t,t')|(h,r,t)\in\Set{G},(h,r,t')\not\in\Set{G}\}$, and $(h,r,t')$ is a broken triplet constructed by replacing one entity in a valid triplet randomly; $\sigma(\cdot)$ is the sigmoid function.
This layer models the entities and relations on the granularity of triples, working as a regularizer and injecting the direct connections into representations, and thus increases the model representation ability (evidences in Section~\ref{sec:ablation-study}.)

\subsection{Attentive Embedding Propagation Layers}
Next we build upon the architecture of graph convolution network~\cite{GCN} to recursively propagate embeddings along high-order connectivity;
moreover, by exploiting the idea of graph attention network~\cite{GAT}, we generate attentive weights of cascaded propagations to reveal the importance of such connectivity.
Here we start by describing a single layer, which consists of three components: \emph{information propagation}, \emph{knowledge-aware attention}, and \emph{information aggregation}, and then discuss how to generalize it to multiple layers.

\vspace{5px}
\noindent\textbf{Information Propagation:}
One entity can be involved in multiple triplets, serving as the bridge connecting two triplets and propagating information.
Taking $e_1\xrightarrow{r_2} i_{2}\xrightarrow{-r_1} u_2$ and $e_2\xrightarrow{r_3} i_{2}\xrightarrow{-r_1} u_2$ as an example, item $i_2$ takes attributes $e_1$ and $e_2$ as inputs to enrich its own features, and then contributes user $u_2$'s preferences, which can be simulated by propagating information from $e_1$ to $u_2$.
We build upon this intuition to perform information propagation between an entity and its neighbors.


Considering an entity $h$, we use $\Set{N}_{h}=\{(h,r,t)|(h,r,t)\in\Set{G}\}$ to denote the set of triplets where $h$ is the head entity, termed ego-network~\cite{DeepInf}.
To characterize the first-order connectivity structure of entity $h$, we compute the linear combination of $h$'s ego-network:
\begin{align}
    \Mat{e}_{\Set{N}_{h}}=\sum_{(h,r,t)\in\Set{N}_{h}}\pi(h,r,t)\Mat{e}_{t},
\end{align}
where $\pi(h,r,t)$ controls the decay factor on each propagation on edge $(h,r,t)$, indicating how much information being propagated from $t$ to $h$ conditioned to relation $r$.



\vspace{5px}
\noindent\textbf{Knowledge-aware Attention:}
We implement $\pi(h,r,t)$ via relational attention mechanism, which is formulated as follows:
\begin{align}\label{equ:attention}
    \pi(h,r,t)=\Trans{(\Mat{W}_{r}\Mat{e}_{t})}\text{tanh}\Big((\Mat{W}_{r}\Mat{e}_{h}+\Mat{e}_{r})\Big),
\end{align}
where we select tanh~\cite{GAT} as the nonlinear activation function.
This makes the attention score dependent on the distance between $e_h$ and $e_t$ in the relation $r$'s space, \eg propagating more information for closer entities.
Note that, we employ only inner product on these representations for simplicity, and leave the further exploration of the attention module as the future work.

Hereafter, we normalize the coefficients across all triplets connected with $h$ by adopting the softmax function:
\begin{align}
    \pi(h,r,t)=\frac{\exp(\pi(h,r,t))}{\sum_{(h,r',t')\in\Set{N}_{h}}\exp(\pi(h,r',t'))}.
\end{align}
As a result, the final attention score is capable of suggesting which neighbor nodes should be given more attention to capture collaborative signals.
When performing propagation forward, the attention flow suggests parts of the data to focus on, which can be treated as explanations behind the recommendation.


Distinct from the information propagation in GCN~\cite{GCN} and GraphSage~\cite{GraphSage} which set the discount factor between two nodes as $1/\sqrt{|\Set{N}_{h}||\Set{N}_{t}|}$ or $1/|\Set{N}_{t}|$, our model not only exploits the proximity structure of graph, but also specify varying importance of neighbors.
Moreover, distinct from graph attention network~\cite{GAT} which only takes node representations as inputs, we model the relation $e_r$ between $e_h$ and $e_t$, encoding more information during propagation.
We perform experiments to verify the effectiveness of the attention mechanism and visualize the attention flow in Section~\ref{sec:ablation-study} and Section~\ref{sec:visualization}, respectively.


\vspace{5px}
\noindent\textbf{Information Aggregation}:
The final phase is to aggregate the entity representation $\Mat{e}_h$ and its ego-network representations $\Mat{e}_{\Set{N}_{h}}$ as the new representation of entity $h$ --- more formally, $\Mat{e}_{h}^{(1)}=f(\Mat{e}_h,\Mat{e}_{\Set{N}_{h}})$.
We implement $f(\cdot)$ using three types of aggregators:
\begin{itemize}[leftmargin=*]
    \item \emph{GCN Aggregator}~\cite{GCN} sums two representations up and applies a nonlinear transformation, as follows:
        \begin{align}
            f_{\text{GCN}}=\text{LeakyReLU}\Big(\Mat{W}(\Mat{e}_{h} + \Mat{e}_{\Set{N}_{h}})\Big),
        \end{align}
        where we set the activation function set as LeakyReLU~\cite{Nonlinear}; $\Mat{W}\in\Space{R}^{d'\times d}$ are the trainable weight matrices to distill useful information for propagation, and $d'$ is the transformation size.
    
    \item \emph{GraphSage Aggregator}~\cite{GraphSage} concatenates two representations, followed by a nonlinear transformation:
        \begin{align}
            f_{\text{GraphSage}}=\text{LeakyReLU}\Big(\Mat{W}(\Mat{e}_{h} || \Mat{e}_{\Set{N}_{h}})\Big),
        \end{align}
        where $||$ is the concatenation operation.
        
    \item \emph{Bi-Interaction Aggregator} is carefully designed by us to consider two kinds of feature interactions between $\Mat{e}_{h}$ and $\Mat{e}_{\Set{N}_{h}}$, as follows:
        \begin{align}
            f_{\text{Bi-Interaction}}=&\text{LeakyReLU}\Big(\Mat{W}_{1}(\Mat{e}_{h} + \Mat{e}_{\Set{N}_{h}})\Big) +\nonumber\\ &\text{LeakyReLU}\Big(\Mat{W}_{2}(\Mat{e}_{h}\odot\Mat{e}_{\Set{N}_{h}})\Big),
        \end{align}
        where $\Mat{W}_1,\Mat{W}_2\in\Space{R}^{d'\times d}$ are the trainable weight matrices, and $\odot$ denotes the element-wise product. Distinct from GCN and GraphSage aggregators, we additionally encode the feature interaction between $\Mat{e}_{h}$ and $\Mat{e}_{\Set{N}_{h}}$. This term makes the information being propagated sensitive to the affinity between $\Mat{e}_{h}$ and $\Mat{e}_{\Set{N}_{h}}$, \eg passing more messages from similar entities.
\end{itemize}
To summarize, the advantage of the embedding propagation layer lies in explicitly exploiting the first-order connectivity information to relate user, item, and knowledge entity representations.
We empirically compare the three aggregators in Section~\ref{sec:aggregator}.

\vspace{5px}
\noindent\textbf{High-order Propagation:}
We can further stack more propagation layers to explore the high-order connectivity information, gathering the information propagated from the higher-hop neighbors.
More formally, in the $l$-th steps, we recursively formulate the representation of an entity as:
\begin{gather}\label{equ:l-aggregator}
    \Mat{e}_{h}^{(l)}=f\Big(\Mat{e}^{(l-1)}_{h}, \Mat{e}^{(l-1)}_{\Set{N}_{h}}\Big),
\end{gather}
wherein the information propagated within $l$-ego network for the entity $h$ is defined as follows,
\begin{gather}\label{equ:l-message}
    \Mat{e}^{(l-1)}_{\Set{N}_{h}}=\sum_{(h,r,t)\in\Set{N}_{h}}\pi(h,r,t)\Mat{e}^{(l-1)}_{t},
\end{gather}
$\Mat{e}_{t}^{(l-1)}$ is the representation of entity $t$ generated from the previous information propagation steps, memorizing the information from its $(l-1)$-hop neighbors;
$\Mat{e}_{h}^{(0)}$ is set as $\Mat{e}_{h}$ at the initial information-propagation iteration.
It further contributes to the representation of entity $h$ at layer $l$.
As a result, high-order connectivity like $u_2\xrightarrow{r_1} i_{2}\xrightarrow{-r_2} e_1 \xrightarrow{r_2} i_1\xrightarrow{-r_1} u_1$ can be captured in the embedding propagation process.
Furthermore, the information from $u_{2}$ is explicitly encoded in $\Mat{e}^{(3)}_{u_{1}}$.
Clearly, the high-order embedding propagation seamlessly injects the attribute-based collaborative signal into the representation learning process.

\subsection{Model Prediction}

After performing $L$ layers, we obtain multiple representations for user node $u$, namely $\{\Mat{e}^{(1)}_{u},\cdots,\Mat{e}^{(L)}_{u}\}$;
analogous to item node $i$, $\{\Mat{e}^{(1)}_{i},\cdots,\Mat{e}^{(L)}_{i}\}$ are obtained.
As the output of the $l$-th layer is the message aggregation of the tree structure depth of $l$ rooted at $u$ (or $i$) as shown in Figure~\ref{fig:intro}, the outputs in different layers emphasize the connectivity information of different orders.
We hence adopt the layer-aggregation mechanism~\cite{JumpKG} to concatenate the representations at each step into a single vector, as follows:
\begin{align}\label{equ:final-rep}
    \Mat{e}^{*}_{u}=\Mat{e}^{(0)}_{u}\Vert\cdots\Vert\Mat{e}^{(L)}_{u},\quad
    \Mat{e}^{*}_{i}=\Mat{e}^{(0)}_{i}\Vert\cdots\Vert\Mat{e}^{(L)}_{i},
\end{align}
where $\Vert$ is the concatenation operation.
By doing so, we not only enrich the initial embeddings by performing the embedding propagation operations, but also allow controlling the strength of propagation by adjusting $L$.

Finally, we conduct inner product of user and item representations, so as to predict their matching score:
\begin{align}
    \hat{y}(u,i)=\Trans{\Mat{e}^{*}_{u}}\Mat{e}^{*}_{i}.
\end{align}

\subsection{Optimization}

To optimize the recommendation model, we opt for the BPR loss~\cite{BPRMF}.
Specifically, it assumes that the observed interactions, which indicate more user preferences, should be assigned higher prediction values than unobserved ones:
\begin{align}\label{equ:cf-loss}
	\Lapl_{\text{CF}}=\sum_{(u,i,j)\in\Set{O}}-\ln\sigma\Big(\hat{y}(u,i)-\hat{y}(u,j)\Big)
\end{align}
where $\Set{O}=\{(u,i,j)|(u,i)\in\Set{R}^{+}, (u,j)\in\Set{R}^{-}\}$ denotes the training set, $\Set{R}^{+}$ indicates the observed (positive) interactions between user $u$ and item $j$ while $\Set{R}^{-}$ is the sampled unobserved (negative) interaction set;
$\sigma(\cdot)$ is the sigmoid function.

Finally, we have the objective function to learn Equations~\eqref{equ:kg-loss} and~\eqref{equ:cf-loss} jointly, as follows:
\begin{align}\label{equ:loss}
    \Lapl_{\text{KGAT}}=\Lapl_{\text{KG}}+\Lapl_{\text{CF}}+\lambda\norm{\Theta}^{2}_{2},
\end{align}
where $\Theta=\{\Mat{E}, \Mat{W}_{r},\forall l\in\Set{R},\Mat{W}_{1}^{(l)},\Mat{W}_{2}^{(l)}, \forall l \in \{1,\cdots,L\}\}$ is the model parameter set, and $\Mat{E}$ is the embedding table for all entities and relations;
$L_{2}$ regularization parameterized by $\lambda$ on $\Theta$ is conducted to prevent overfitting.
It is worth pointing out that in terms of model size, the majority of model parameters comes from the entity embeddings (\eg 6.5 million on experimented Amazon dataset), which is almost identical to that of FM;
the propagation layer weights are lightweight (\eg 5.4 thousand for the tower structure of three layers, \ie $64-32-16-8$, on the Amazon dataset).

\subsubsection{\textbf{Training:}}
We optimize $\Lapl_{KG}$ and $\Lapl_{CF}$ alternatively, where mini-batch Adam~\cite{Adam} is adopted to optimize the embedding loss and the prediction loss.
Adam is a widely used optimizer, which is able to adaptively control the learning rate \wrt the absolute value of gradient.
In particular, for a batch of randomly sampled $(h,r,t,t')$, we update the embeddings for all nodes; hereafter, we sample a batch of $(u,i,j)$ randomly, retrieve their representations after $L$ steps of propagation, and then update model parameters by using the gradients of the prediction loss.

\subsubsection{\textbf{Time Complexity Analysis:}}
As we adopt the alternative optimization strategy, the time cost mainly comes from two parts.
For the knowledge graph embedding (\cf Equation~\eqref{equ:kg-loss}), the translation principle has computational complexity $O(|\Set{G}_{2}|d^{2})$.
For the attention embedding propagation part, the matrix multiplication of the $l$-th layer has computational complexity $O(|\Set{G}|d_{l}d_{l-1})$; and $d_{l}$ and $d_{l-1}$ are the current and previous transformation size.
For the final prediction layer, only the inner product is conducted, for which the time cost of the whole training epoch is $O(\sum_{l=1}^{L}|\Set{G}|d_{l})$.
Finally, the overall training complexity of KGAT is $O(|\Set{G}_{2}|d^{2}+\sum_{l=1}^{L}|\Set{G}|d_{l}d_{l-1} +|\Set{G}|d_{l})$.

As online services usually require real-time recommendation, the computational cost during inference is more important that that of training phase. Empirically, FM, NFM, CFKG, CKE, GC-MC, \textbf{KGAT}, MCRec, and RippleNet cost around $700$s, $780$s, $800$s, $420$s, $500$s, $\Mat{560}$s, $20$ hours, and $2$ hours for all testing instances on Amazon-Book dataset, respectively.
As we can see, KGAT achieves comparable computation complexity to SL models (FM and NFM) and regularization-based methods (CFKG and CKE), being much efficient that path-based methods (MCRec and RippleNet).

\section{Experiments}

We evaluate our proposed method, especially the embedding propagation layer, on three real-world datasets.
We aim to answer the following research questions:
\begin{itemize}[leftmargin=*]
	\item \textbf{RQ1}: How does KGAT perform compared with state-of-the-art knowledge-aware recommendation methods?
	\item \textbf{RQ2}: How do different components (\ie knowledge graph embedding, attention mechanism, and aggregator selection) affect KGAT?
	\item \textbf{RQ3}: Can KGAT provide reasonable explanations about user preferences towards items?
\end{itemize}

\subsection{Dataset Description}

To evaluate the effectiveness of KGAT, we utilize three benchmark datasets: Amazon-book, Last-FM, and Yelp2018, which are publicly accessible and vary in terms of domain, size, and sparsity.

\vspace{2px}
\noindent\textbf{Amazon-book\footnote{\url{http://jmcauley.ucsd.edu/data/amazon}.}:}
Amazon-review is a widely used dataset for product recommendation~\cite{amazon-review}.
We select Amazon-book from this collection.
To ensure the quality of the dataset, we use the $10$-core setting, \ie retaining users and items with at least ten interactions.

\vspace{2px}
\noindent\textbf{Last-FM\footnote{\url{https://grouplens.org/datasets/hetrec-2011/}.}:}
This is the music listening dataset collected from Last.fm online music systems.
Wherein, the tracks are viewed as the items.
In particular, we take the subset of the dataset where the timestamp is from Jan, 2015 to June, 2015.
We use the same $10$-core setting in order to ensure data quality.

\vspace{2px}
\noindent\textbf{Yelp2018\footnote{\url{https://www.yelp.com/dataset/challenge}.}:}
This dataset is adopted from the 2018 edition of the Yelp challenge.
Here we view the local businesses like restaurants and bars as the items.
Similarly, we use the $10$-core setting to ensure that each user and item have at least ten interactions.

Besides the user-item interactions, we need to construct item knowledge for each dataset.
For Amazon-book and Last-FM, we follow the way in~\cite{KB4Rec} to map items into Freebase entities via title matching if there is a mapping available.
In particular, we consider the triplets that are directly related to the entities aligned with items, no matter which role (\ie subject or object) it serves as.
Distinct from existing knowledge-aware datasets that provide only one-hop entities of items, we also take the triplets that involve two-hop neighbor entities of items into consideration.
For Yelp2018, we extract item knowledge from the local business information network (\eg category, location, and attribute) as KG data.
To ensure the KG quality, we then preprocess the three KG parts by filtering out infrequent entities (\ie lowever than $10$ in both datasets) and retaining the relations appearing in at least $50$ triplets.
We summarize the statistics of three datasets in Table~\ref{tab:dataset}.

For each dataset, we randomly select $80\%$ of interaction history of each user to constitute the training set, and treat the remaining as the test set.
From the training set, we randomly select $10\%$ of interactions as validation set to tune hyper-parameters.
For each observed user-item interaction, we treat it as a positive instance, and then conduct the negative sampling strategy to pair it with one negative item that the user did not consume before.

\subsection{Experimental Settings}

\begin{table}[t]
\caption{Statistics of the datasets.}
\vspace{-10px}
\label{tab:dataset}
\resizebox{0.46\textwidth}{!}{
\begin{tabular}{c l|r|r|r}
\hline
 &  & \multicolumn{1}{l|}{Amazon-book} & \multicolumn{1}{l|}{Last-FM} & \multicolumn{1}{l}{Yelp2018} \\ \hline\hline
\multirow{3}{*}{\begin{tabular}[c]{@{}c@{}}User-Item\\ Interaction\end{tabular}} & \#Users & $70,679$ & $23,566$ & $45,919$ \\
 & \#Items & $24,915$ & $48,123$ & $45,538$ \\
 & \#Interactions & $847,733$ & $3,034,796$ & $1,185,068$ \\ \hline\hline
\multirow{3}{*}{\begin{tabular}[c]{@{}c@{}}Knowledge\\ Graph\end{tabular}} & \#Entities & $88,572$ & $58,266$ & $90,961$ \\
 & \#Relations & $39$ & $9$ & $42$ \\ 
 & \#Triplets & $2,557,746$ & $464,567$ & $1,853,704$ \\ \hline
\end{tabular}}
\vspace{-15px}
\end{table}

\subsubsection{\textbf{Evaluation Metrics}}
For each user in the test set, we treat all the items that the user has not interacted with as the negative items.
Then each method outputs the user's preference scores over all the items, except the positive ones in the training set.
To evaluate the effectiveness of top-$K$ recommendation and preference ranking, we adopt two widely-used evaluation protocols~\cite{NCF,HOP-rec}: recall@$K$ and ndcg@$K$.
By default, we set $K=20$.
We report the average metrics for all users in the test set.

\subsubsection{\textbf{Baselines}}
To demonstrate the effectiveness, we compare our proposed KGAT with SL (FM and NFM), regularization-based (CFKG and CKE), path-based (MCRec and RippleNet), and graph neural network-based (GC-MC) methods, as follows:

\begin{itemize}[leftmargin=*]
	\item \textbf{FM}~\cite{FM}:~This is a bechmark factorization model, where considers the second-order feature interactions between inputs. Here we treat IDs of a user, an item, and its knowledge (\ie entities connected to it) as input features.
	
	\item \textbf{NFM}~\cite{NFM}: The method is a state-of-the-art factorization model, which subsumes FM under neural network. Specially, we employed one hidden layer on input features as suggested in~\cite{NFM}.
	
	\item \textbf{CKE}~\cite{CKE}: This is a representative regularization-based method, which exploits semantic embeddings derived from TransR~\cite{TransR} to enhance matrix factorization~\cite{BPRMF}.
    
    \item \textbf{CFKG}~\cite{CFKG}: The model applies TransE~\cite{TransE} on the unified graph including users, items, entities, and relations, casting the recommendation task as the plausibility prediction of $(u,\emph{Interact},i)$ triplets.
    
    \item \textbf{MCRec}~\cite{MCRec}: This is a path-based model, which extracts qualified meta-paths as connectivity between a user and an item.
    
	\item \textbf{RippleNet}~\cite{RippleNet}: Such model combines regularization- and path-based methods, which enrich user representations by adding that of items within paths rooted at each user.
	
	\item \textbf{GC-MC}~\cite{GC-MC}: Such model is designed to employ GCN~\cite{GCN} encoder on graph-structured data, especially for the user-item bipartite graph. Here we apply it on the user-item knowledge graph. Especially, we employ one graph convolution layers as suggested in~\cite{GC-MC}, where the hidden dimension is set equal to the embedding size.
\end{itemize}

\subsubsection{\textbf{Parameter Settings}}\label{sec:experiment-settings}
We implement our KGAT model in Tensorflow.
The embedding size is fixed to $64$ for all models, except RippleNet $16$ due to its high computational cost.
We optimize all models with Adam optimizer, where the batch size is fixed at $1024$.
The default Xavier initializer~\cite{Xarvier} to initialize the model parameters.
We apply a grid search for hyper-parameters: the learning rate is tuned amongst $\{0.05, 0.01, 0.005, 0.001\}$, the coefficient of $L_2$ normalization is searched in $\{10^{-5},10^{-4},\cdots,10^{1},10^{2}\}$, and the dropout ratio is tuned in $\{0.0,0.1,\cdots,0.8\}$ for NFM, GC-MC, and KGAT.
Besides, we employ the node dropout technique for GC-MC and KGAT, where the ratio is searched in $\{0.0,0.1,\cdots,0.8\}$.
For MCRec, we manually define several types of user-item-attribute-item meta-paths, such as \emph{user-book-author-user} and \emph{user-book-genre-user} for Amazon-book dataset; we set the hidden layers as suggested in~\cite{MCRec}, which is a tower structure with $512$, $256$, $128$, $64$ dimensions.
For RippleNet, we set the number of hops and the memory size as $2$ and $8$, respectively.
Moreover, early stopping strategy is performed, \ie premature stopping if recall@$20$ on the validation set does not increase for $50$ successive epochs.
To model the third-order connectivity, we set the depth of KGAT $L$ as three with hidden dimension $64$, $32$, and $16$, respectively; we also report the effect of layer depth in Section~\ref{sec:layer-depth}.
For each layer, we conduct the Bi-Interaction aggregator.

\subsection{Performance Comparison (RQ1)}
We first report the performance of all the methods, and then investigate how the modeling of high-order connectivity alleviate the sparsity issues.

\subsubsection{\textbf{Overall Comparison}}

\begin{figure*}[t]
\centering
\subfigure[ndcg on Amazon-Book]{
\label{fig:sparsity-amazon-book}\includegraphics[width=0.32\textwidth]{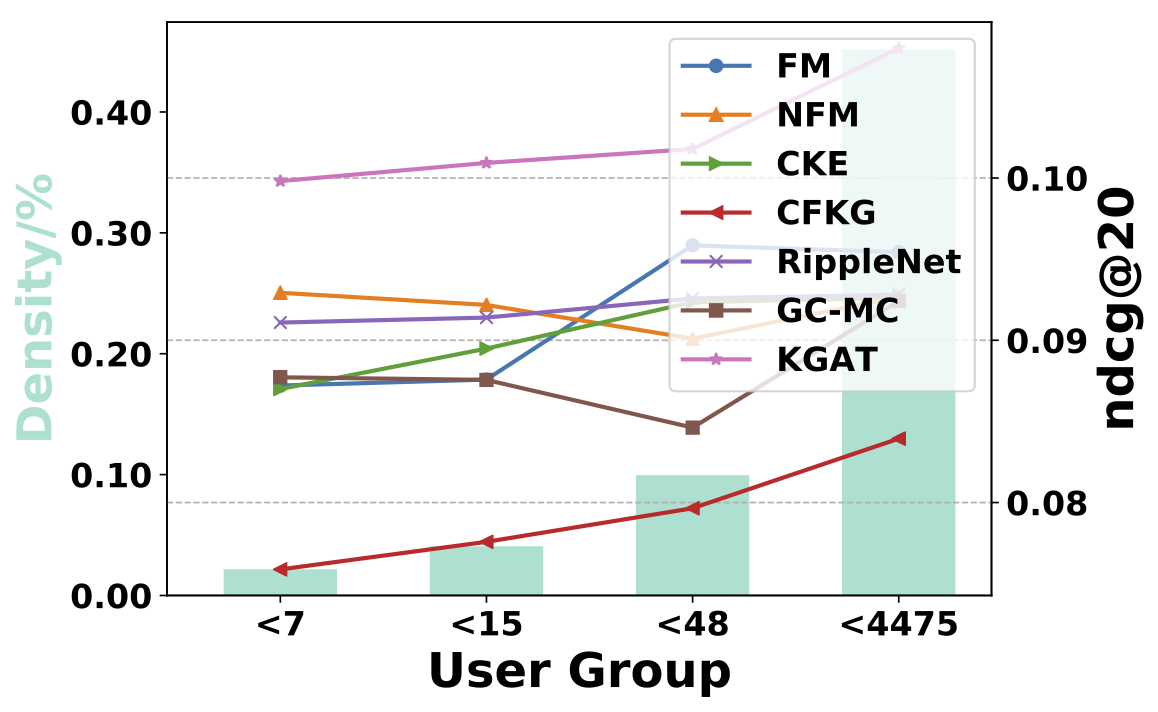}}
\subfigure[ndcg on Last-FM]{
\label{fig:sparsity-last-fm}\includegraphics[width=0.32\textwidth]{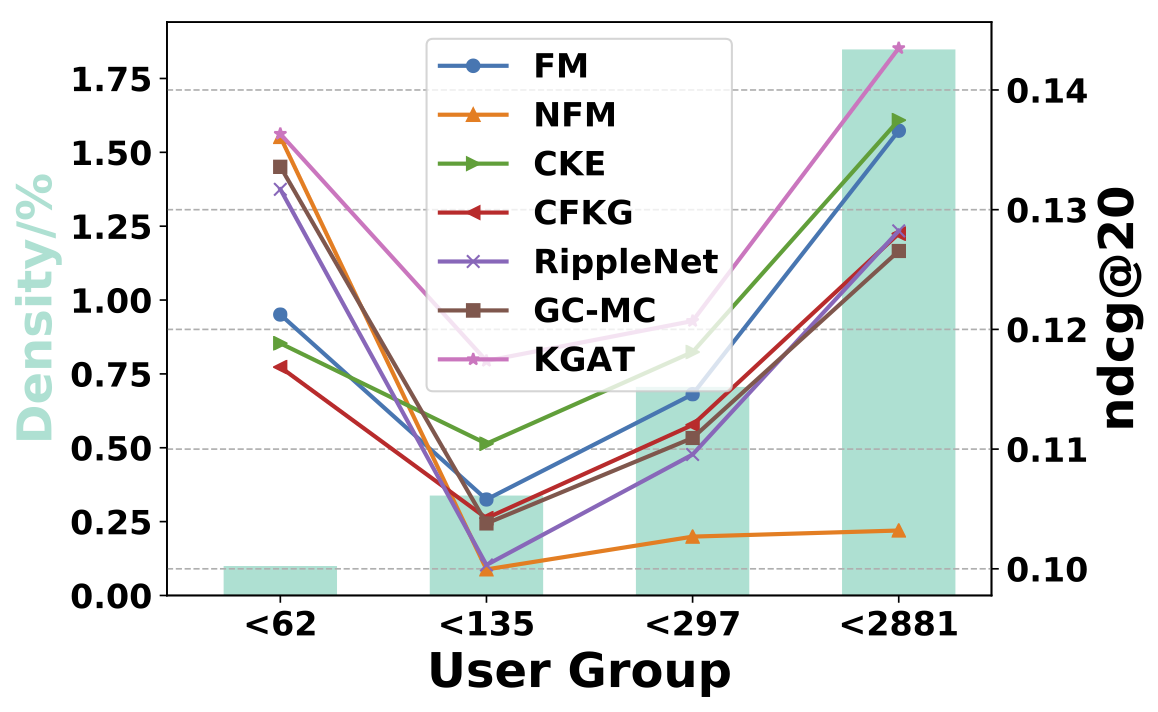}}
\subfigure[ndcg on Yelp2018]{
\label{fig:sparsity-yelp2018}\includegraphics[width=0.32\textwidth]{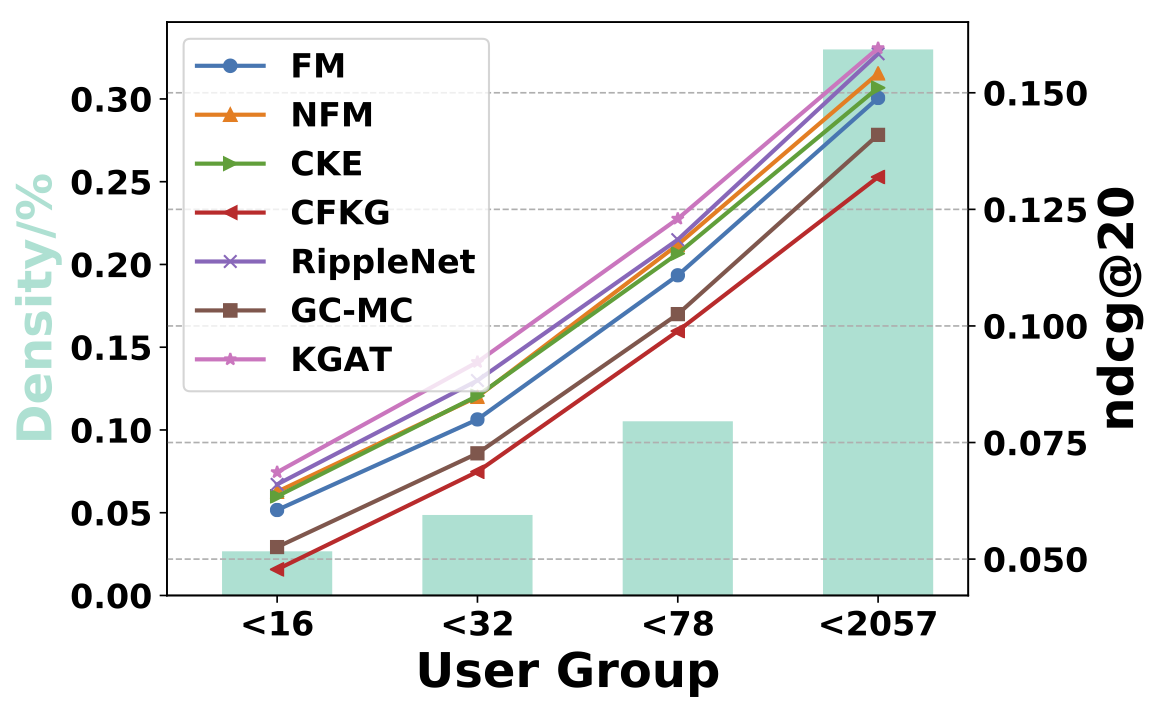}}
\vspace{-15pt}
\caption{Performance comparison over the sparsity distribution of user groups on different datasets.
The background histograms indicate the density of each user group; meanwhile, the lines demonstrate the performance \wrt ndcg@$20$.}
\label{fig:sparsity}\vspace{-10pt}
\end{figure*}

\begin{table}[t]
\caption{Overall Performance Comparison.}
\vspace{-10px}
\label{tab:overall-performance}
\resizebox{0.48\textwidth}{!}{
\begin{tabular}{l|c c | c c | c c}
\hline
 & \multicolumn{2}{c|}{Amazon-Book} & \multicolumn{2}{c|}{Last-FM} & \multicolumn{2}{c}{Yelp2018} \\
 & recall & ndcg & recall & ndcg & recall & ndcg \\ \hline\hline
FM & $0.1345$ & $0.0886$ & $0.0778$ & $0.1181$ & $0.0627$ & $0.0768$ \\
NFM & $\Mat{0.1366}$ & $0.0913$ & $\Mat{0.0829}$ & $0.1214$ & $0.0660$ & $0.0810$ \\ \hline
CKE & $0.1343$ & $0.0885$ & $0.0736$ & $0.1184$ & $0.0657$ & $0.0805$ \\
CFKG & $0.1142$ & $0.0770$ & $0.0723$ & $0.1143$ & $0.0522$ & $0.0644$ \\ \hline
MCRec & $0.1113$ & $0.0783$ & - & - & - & - \\
RippleNet & $0.1336$ & $0.0910$ & $0.0791$ & $0.1238$ & $\Mat{0.0664}$ & $\Mat{0.0822}$ \\
GC-MC & $0.1316$ & $0.0874$ & $0.0818$ & $\Mat{0.1253}$ & $0.0659$ & $0.0790$ \\ \hline
KGAT & $\Mat{0.1489^{*}}$ & $\Mat{0.1006^{*}}$ & $\Mat{0.0870^{*}}$ & $\Mat{0.1325^{*}}$ & $\Mat{0.0712^{*}}$ & $\Mat{0.0867^{*}}$ \\ \hline\hline
\%Improv. & $8.95\%$ & $10.05\%$ & $4.93\%$ & $5.77\%$ & $7.18\%$ & $5.54\%$ \\ \hline
\end{tabular}}
\vspace{-15px}
\end{table}

The performance comparison results are presented in Table~\ref{tab:overall-performance}. We have the following observations:
\begin{itemize}[leftmargin=*]
    \item KGAT consistently yields the best performance on all the datasets. In particular, KGAT improves over the strongest baselines \wrt recall@$20$ by $8.95\%$, $4.93\%$, and $7.18\%$ in Amazon-book, Last-FM, and Yelp2018, respectively.
    By stacking multiple attentive embedding propagation layers, KGAT is capable of exploring the high-order connectivity in an explicit way, so as to capture collaborative signal effectively.
    This verifies the significance of capturing collaborative signal to transfer knowledge.
    Moreover, compared with GC-MC, KGAT justifies the effectiveness of the attention mechanism, specifying the attentive weights \wrt compositional semantic relations, rather than the fixed weights used in GC-MC.
    
    \item SL methods (\ie FM and NFM) achieve better performance than the CFKG and CKE in most cases, indicating that regularization-based methods might not make full use of item knowledge. In particular, to enrich the representation of an item, FM and NFM exploit the embeddings of its connected entities, while CFKG and CKE only use that of its aligned entities. Furthermore, the cross features in FM and NFM actually serve as the second-order connectivity between users and entities, whereas CFKG and CKE model connectivity on the granularity of triples, leaving high-order connectivity untouched.
    
    
    \item Compared to FM, the performance of RippleNet verifies that incorporating two-hop neighboring items is of importance to enrich user representations. It therefore points to the positive effect of modeling the high-order connectivity or neighbors. However, RippleNet slightly underperforms NFM in Amazon-book and Last-FM, while performing better in Yelp2018.
    One possible reason is that NFM has stronger expressiveness, since the hidden layer allows NFM to capture the nonlinear and complex feature interactions between user, item, and entity embeddings.
    
    \item RippleNet outperforms MCRec by a large margin in Amazon-book. One possible reason is that MCRec depends heavily on the quality of meta-paths, which require extensive domain knowledge to define. The observation is consist with~\cite{RippleNet}.
    
    \item GC-MC achieves comparable performance to RippleNet in Last-FM and Yelp2018 datasets. While introducing the high-order connectivity into user and item representations, GC-MC forgoes the semantic relations between nodes; whereas RippleNet utilizes relations to guide the exploration of user preferences.

\end{itemize}

\subsubsection{\textbf{Performance Comparison \wrt Interaction Sparsity Levels}}\label{sec:sparsity}
One motivation to exploiting KG is to alleviate the sparsity issue, which usually limits the expressiveness of recommender systems.
It is hard to establish optimal representations for inactive users with few interactions.
Here we investigate whether exploiting connectivity information helps alleviate this issue.


Towards this end, we perform experiments over user groups of different sparsity levels.
In particular, we divide the test set into four groups based on interaction number per user, meanwhile try to keep different groups have the same total interactions.
Taking Amazon-book dataset as an example, the interaction numbers per user are less than $7$, $15$, $48$, and $4475$ respectively.
Figure~\ref{fig:sparsity} illustrates the results \wrt ndcg@$20$ on different user groups in Amazon-book, Last-FM, and Yelp2018.
We can see that:
\begin{itemize}[leftmargin=*]
    \item KGAT outperforms the other models in most cases, especially on the two sparsest user groups in Amazon-Book and Yelp2018.
    It again verifies the significance of high-order connectivity modeling, which 1) contains the lower-order connectivity used in baselines, and 2) enriches the representations of inactive users via recursive embedding propagation.

    
    
    \item It is worthwhile pointing out that KGAT slightly outperforms some baselines in the densest user group (\eg the $<2057$ group of Yelp2018). One possible reason is that the preferences of users with too many interactions are too general to capture. High-order connectivity could introduce more noise into the user preferences, thus leading to the negative effect.
\end{itemize}

\subsection{Study of KGAT (RQ2)}

\begin{table}[t]
\caption{Effect of embedding propagation layer numbers ($L$).
}
\vspace{-10px}
\label{tab:depth}
\resizebox{0.48\textwidth}{!}{
\begin{tabular}{l|c c|c c|c c}
\hline
 & \multicolumn{2}{c|}{Amazon-Book} & \multicolumn{2}{c|}{Last-FM} & \multicolumn{2}{c}{Yelp2018} \\ 
 & recall & ndcg & recall & ndcg & recall & ndcg \\ \hline\hline
KGAT-1 & 0.1393 & 0.0948 & 0.0834 & 0.1286 & 0.0693 & 0.0848 \\ 
KGAT-2 & 0.1464 & 0.1002 & 0.0863 & 0.1318 & 0.0714 & \textbf{0.0872} \\ 
KGAT-3 & 0.1489 & 0.1006 & 0.0870 & 0.1325 & 0.0712 & 0.0867 \\ 
KGAT-4 & \textbf{0.1503} & \textbf{0.1015} & \textbf{0.0871} & \textbf{0.1329} & \textbf{0.0722} & 0.0871 \\ \hline
\end{tabular}}
\vspace{-10px}
\end{table}

To get deep insights on the attentive embedding propagation layer of KGAT, we investigate its impact.
We first study the influence of layer numbers.
In what follows, we explore how different aggregators affect the performance.
We then examine the influence of knowledge graph embedding and attention mechanism.

\subsubsection{\textbf{Effect of Model Depth}}\label{sec:layer-depth}
We vary the depth of KGAT (\eg $L$) to investigate the efficiency of usage of multiple embedding propagation layers.
In particular, the layer number is searched in the range of $\{1,2,3,4\}$; we use KGAT-1 to indicate the model with one layer, and similar notations for others.
We summarize the results in Table~\ref{tab:depth}, and have the following observations:
\begin{itemize}[leftmargin=*]
    \item Increasing the depth of KGAT is capable of boosting the performance substantially. Clearly, KGAT-2 and KGAT-3 achieve consistent improvement over KGAT-1 across all the board. We attribute the improvements to the effective modeling of high-order relation between users, items, and entities, which is carried by the second- and third-order connectivities, respectively.
    
    \item Further stacking one more layer over KGAT-3, we observe that KGAT-4 only achieve marginal improvements. It suggests that considering third-order relations among entities could be sufficient to capture the collaborative signal, which is consistent to the findings in~\cite{KGreasoning19,MCRec}.
    
    \item Jointly analyzing Tables~\ref{tab:overall-performance} and~\ref{tab:depth}, KGAT-1 consistently outperforms other baselines in most cases. It again verifies the effectiveness of that attentive embedding propagation, empirically showing that it models the first-order relation better.
\end{itemize}

\subsubsection{\textbf{Effect of Aggregators}}\label{sec:aggregator}

To explore the impact of aggregators, we consider the variants of KGAT-1 that uses different settings --- more specifically GCN, GraphSage, and Bi-Interaction (\cf Section~~\ref{sec:embedding-propagation}), termed KGAT-1$_{\text{GCN}}$, KGAT-1$_{\text{GraphSage}}$, and KGAT-1$_{\text{Bi}}$, respectively.
Table~\ref{tab:aggregator} summarizes the experimental results.
We have the following findings:
\begin{itemize}[leftmargin=*]
    \item KGAT-1$_{\text{GCN}}$ is consistently superior to KGAT-1$_{\text{GraphSage}}$. One possible reason is that GraphSage forgoes the interaction between the entity representation $\Mat{e}_{h}$ and its ego-network representation $\Mat{e}_{\Set{N}_{h}}$. It hence illustrates the importance of feature interaction when performing information aggregation and propagation.
    
    \item Compared to KGAT-1$_{\text{GCN}}$, the performance of KGAT-1$_{\text{Bi}}$ verifies that incorporating additional feature interaction can improve the representation learning. It again illustrates the rationality and effectiveness of Bi-Interaction aggregator.
\end{itemize}

\subsubsection{\textbf{Effect of Knowledge Graph Embedding and Attention Mechanism}}\label{sec:ablation-study}


\begin{table}[t]
\caption{Effect of aggregators.}
\vspace{-10px}
\label{tab:aggregator}
\resizebox{0.48\textwidth}{!}{
\begin{tabular}{l|c c|c c|c c}
\hline
 & \multicolumn{2}{c|}{Amazon-Book} & \multicolumn{2}{c|}{Last-FM} & \multicolumn{2}{c}{Yelp2018} \\ 
Aggregator & recall & ndcg & recall & ndcg & recall & ndcg \\ \hline\hline
GCN & 0.1381 & 0.0931 & 0.0824 & 0.1278 & 0.0688 & 0.0847 \\ 
GraphSage & 0.1372 & 0.0929 & 0.0822 & 0.1268 & 0.0666 & 0.0831 \\ 
Bi-Interaction & \textbf{0.1393} & \textbf{0.0948} & \textbf{0.0834} & \textbf{0.1286} & \textbf{0.0693} & \textbf{0.0848} \\ \hline
\end{tabular}}
\vspace{-10px}
\end{table}

\begin{table}[t]
\caption{Effect of knowledge graph embedding and attention mechanism.
}
\vspace{-10px}
\label{tab:att&kge}
\resizebox{0.48\textwidth}{!}{
\begin{tabular}{l|c c|c c|c c}
\hline
 & \multicolumn{2}{c|}{Amazon-Book} & \multicolumn{2}{c|}{Last-FM} & \multicolumn{2}{c}{Yelp2018} \\ 
 & recall & ndcg & recall & ndcg & recall & ndcg \\ \hline\hline
w/o K\&A & 0.1367 & 0.0928 & 0.0819 & 0.1252 & 0.0654 & 0.0808 \\ 
w/o KGE  & 0.1380 & 0.0933 & 0.0826 & 0.1273 & 0.0664 & 0.0824 \\ 
w/o Att  & 0.1377 & 0.0930 & 0.0826 & 0.1270 & 0.0657 & 0.0815 \\ \hline
\end{tabular}}
\vspace{-10px}
\end{table}

\begin{figure}[t]
    \centering
	\includegraphics[width=0.44\textwidth]{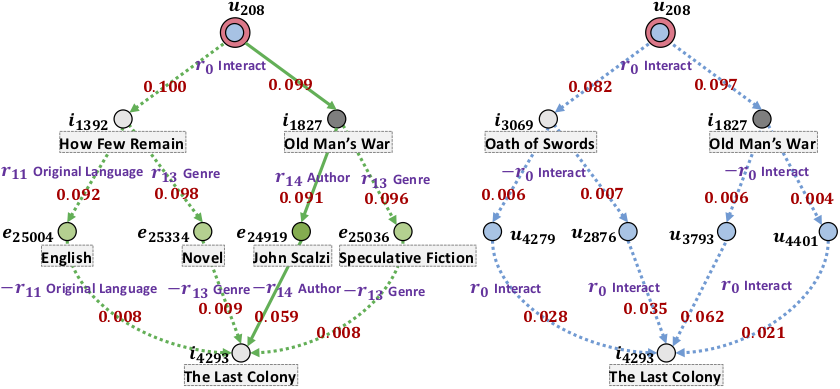}
	\vspace{-10pt}
	\caption{Real Example from Amazon-Book.}
	\label{fig:case-study}
	\vspace{-15pt}
\end{figure}

To verify the impact of knowledge graph embedding and attention mechanism, we do ablation study by considering three variants of KGAT-1.
In particular, we disable the TransR embedding component (\cf Equation~\eqref{equ:kg-loss}) of KGAT, termed KGAT-1$_{\text{w/o KGE}}$;
we disable the attention mechanism (\cf Equation~\eqref{equ:attention}) and set $\pi(h,r,t)$ as $1/|\Set{N}_{h}|$, termed KGAT-1$_{\text{w/o Att}}$.
Moreover, we obtain another variant by removing both components, named KGAT-1$_{\text{w/o K\&A}}$.
We summarize the experimental results in Table~\ref{tab:att&kge} and have the following findings:
\begin{itemize}[leftmargin=*]
    \item Removing knowledge graph embedding and attention components degrades the model's performance. KGAT-1$_{\text{w/o K\&A}}$ consistently underperforms KGAT-1$_{\text{w/o KGE}}$ and KGAT-1$_{\text{w/o Att}}$.
    It makes sense since KGAT$_{\text{w/o K\&A}}$ fails to explicitly model the representation relatedness on the granularity of triplets.

    \item Compared with KGAT-1$_{\text{w/o Att}}$, KGAT-1$_{\text{w/o KGE}}$ performs better in most cases. One possible reason is that treating all neighbors equally (\ie KGAT-1$_{\text{w/o Att}}$) might introduce noises and mislead the embedding propagation process. 
    It verifies the substantial influence of graph attention mechanism.
\end{itemize}

\subsection{Case Study (RQ3)}\label{sec:visualization}
Benefiting from the attention mechanism, we can reason on high-order connectivity to infer the user preferences on the target item, offering explanations.
Towards this end, we randomly selected one user $u_{208}$ from Amazon-Book, and one relevant item $i_{4293}$ (from the test, unseen in the training phase).
We extract behavior-based and attribute-based high-order connectivity connecting the user-item pair, based on the attention scores.
Figure~\ref{fig:case-study} shows the visualization of high-order connectivity. There are two key observations:
\begin{itemize}[leftmargin=*]
    \item KGAT captures the behavior-based and attribute-based high-order connectivity, which play a key role to infer user preferences.
    The retrieved paths can be viewed as the evidence why the item meets the user's preference. As we can see, the connectivity $u_{208}\xrightarrow{r_0} \emph{Old Man's War}\xrightarrow{r_{14}} \emph{John Scalzi}\xrightarrow{-r_{14}} i_{4293}$ has the highest attention score, labeled with the solid line in the left subfigure.
    Hence, we can generate the explanation as \emph{The Last Colony is recommended since you have watched Old Man's War written by the same author John Scalzi}.
    
    \item The quality of item knowledge is of crucial importance. As we can see, entity \emph{English} with relation \emph{Original Language} is involved in one path, which is too general to provide high-quality explanations. This inspires us to perform hard attention to filter less informative entities out in future work. 
\end{itemize}

\section{Conclusion and Future Work}

In this work, we explore high-order connectivity with semantic relations in CKG for knowledge-aware recommendation.
We devised a new framework KGAT, which explicitly models the high-order connectivities in CKG in an end-to-end fashion.
At it core is the attentive embedding propagation layer, which adaptively propagates the embeddings from a node's neighbors to update the node's representation.
Extensive experiments on three real-world datasets demonstrate the rationality and effectiveness of KGAT.

This work explores the potential of graph neural networks in recommendation, and represents an initial attempt to exploit structural knowledge with information propagation mechanism.
Besides knowledge graph, many other structural information indeed exists in real-world scenarios, such as social networks and item contexts.
For example, by integrating social network with CKG, we can investigate how social influence affects the recommendation.
Another exciting direction is the integration of information propagation and decision process, which opens up research possibilities of explainable recommendation.

\vspace{1px}
\noindent\textbf{Acknowledgement:}
This research is part of NExT++ research and also supported by the Thousand Youth Talents Program 2018.
NExT++ is supported by the National Research Foundation, Prime Minister's Office, Singapore under its IRC@SG Funding Initiative.




\bibliographystyle{ACM-Reference-Format}
\balance

\bibliography{ms}
\balance

\end{document}